\documentclass[letterpaper]{article} 
\usepackage[preprint]{aaai2027}  

\usepackage[hyphens]{url}  
\usepackage{graphicx} 
\usepackage{natbib}  
\usepackage{caption} 
\usepackage{amsmath}
\usepackage{amssymb}
\usepackage{booktabs}
\usepackage{multirow}
\usepackage{array}

\title{SkillTV-Bench: Benchmarking How Well Judges Perform \\ on Skill-Augmented Agentic Execution}

\author{
    Zhi Han\textsuperscript{\rm 1,2},
    Chenxi Zeng\textsuperscript{\rm 2},
    Liuhaichen Yang\textsuperscript{\rm 2,4},
    Zihan Guo\textsuperscript{\rm 2,3$^{*}$},
    Ming Zhou\textsuperscript{\rm 2$^{*}$},
    Yang Li\textsuperscript{\rm 1,2$^{*}$}
}

\affiliations{
    \textsuperscript{1} Shanghai Jiao Tong University,
    \textsuperscript{2} Shanghai Artificial Intelligence Laboratory,\\
    \textsuperscript{3} Sun Yat-Sen University,
    \textsuperscript{4} University College London\\
    $^{*}$~Corresponding author.\\
    hanzhi\_sjtu669@sjtu.edu.cn
}

\newcommand{\method}{SkillTV-Evolve}
\newcommand{\dataset}{SkillTV-Bench}

\begin{document}

\maketitle

\begin{abstract}

LLM agents increasingly execute long-horizon tasks through tool use and environment interaction, shifting evaluation from final-response scoring to verification of complete executions. For skill-augmented agents, verification additionally requires the procedural knowledge encoded in task-time skills, because this knowledge indicates what evidence to inspect and which failures are task-critical. However, existing judge benchmarks often expose final responses or static trajectories, and rarely  combine task-time skills with directly inspectable artifacts and environments. We therefore introduce SkillTV-Bench, a 681-case benchmark of real agent trajectories from 50 tasks across eleven domains, designed to evaluate skill-aware trajectory verification for both LLM-as-a-Judge and Agent-as-a-Judge methods. Additionally, we propose SkillTV-Evolve, which externalizes verification knowledge as a reusable JudgeSkill that guides an agent judge to plan targeted inspections and issue evidence-grounded verdicts. On a disjoint development pool, an automated evolution loop further refines the JudgeSkill using misjudged cases. On SkillTV-Bench, the refined skill increases the same agent judge’s accuracy by 14.8 percentage points. In offline rollout-pool selection, it increases selected-trajectory success from 22.9\% with one rollout to 45.5\% with ten rollouts. The code and data are available at \url{https://github.com/HanZhi306/SkillTV-Bench}.
\end{abstract}

\begin{figure}[!t]
    \centering
    \includegraphics[width=\columnwidth]{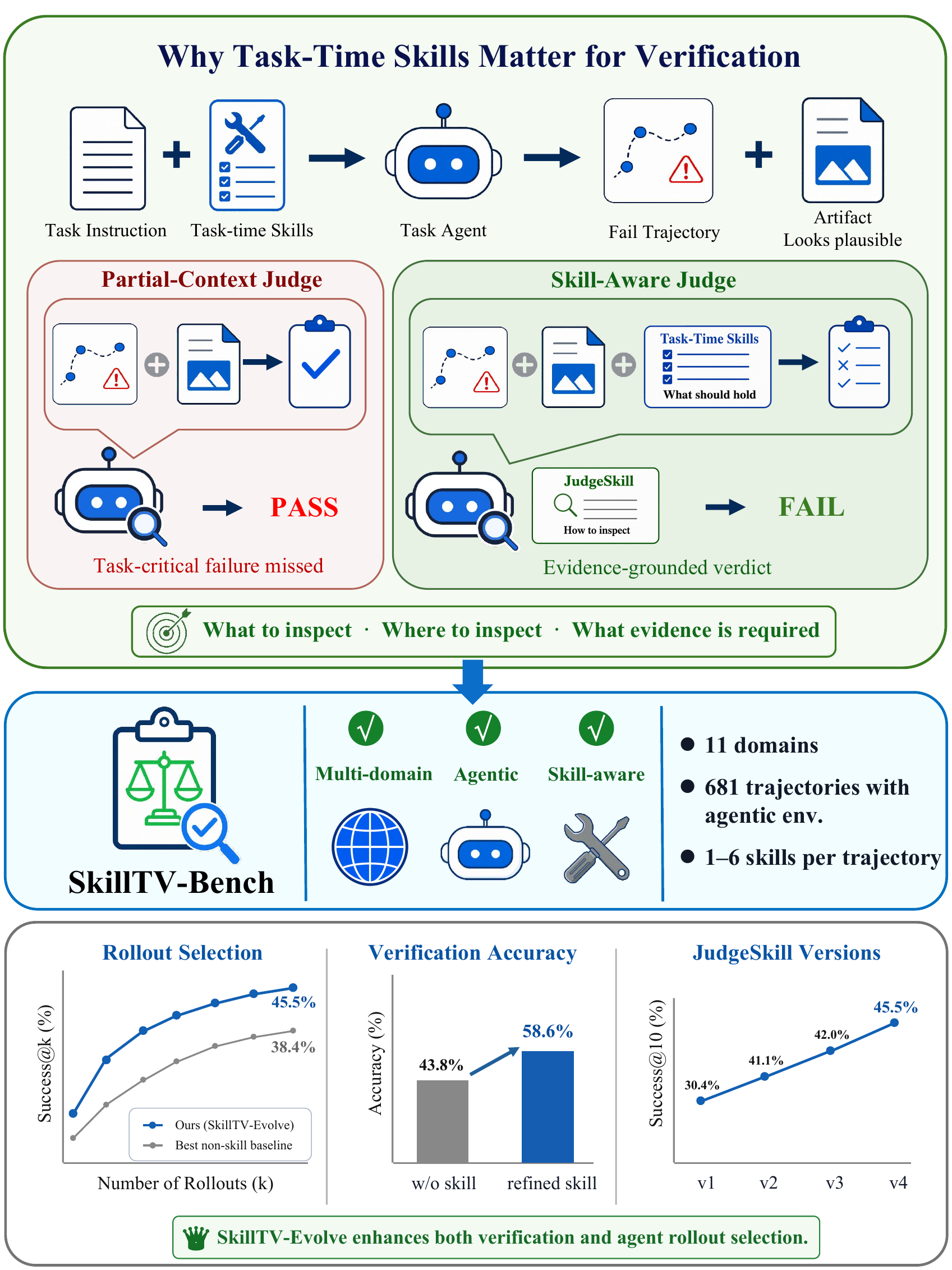}
    \caption{Overview of SkillTV-Bench and SkillTV-Evolve.
    SkillTV-Bench expands domain coverage, agentic interaction,
    and skill awareness beyond existing benchmarks, while SkillTV-Evolve improves rollout selection and trajectory verification
    through iterative JudgeSkill evolution.}
    \label{fig:overview}
\end{figure}

\begin{table*}[t]
\centering
\begin{tabular}{
@{}
>{\raggedright\arraybackslash}p{0.33\textwidth}
c
>{\raggedright\arraybackslash}p{0.19\textwidth}
ccc
@{}
}
\toprule
\multirow{2}{*}[-0.1ex]{\textbf{Benchmark}}
& \multicolumn{1}{c}{\textbf{Evaluation}}
& \multicolumn{1}{c}{\textbf{Evaluation}}
& \textbf{Multi-}
& \textbf{Skill-}
& \textbf{Agentic} \\

& \multicolumn{1}{c}{\textbf{Target}}
& \multicolumn{1}{c}{\textbf{Input}}
& \textbf{Domain}
& \textbf{Aware}
& \textbf{Interaction} \\
\midrule

RewardBench~\cite{lambert2024rewardbenchevaluatingrewardmodels}
& LLM
& Question--answer pairs
& $\checkmark$
& $\times$
& $\times$ \\

JudgeBench~\cite{tan2025judgebenchbenchmarkevaluatingllmbased}
& LLM
& Question--answer pairs
& $\checkmark$
& $\times$
& $\times$\\

AgentRewardBench~\cite{lu2025agentrewardbenchevaluatingautomaticevaluations}
& LLM
& Web-agent trajectories
& $\checkmark$
& $\times$
& $\times$ \\

DevAI~\cite{zhuge2024agentasajudgeevaluateagentsagents}
& Agent
& Code-generation tasks
& $\times$
& $\times$
& $\checkmark$ \\

Plan-RewardBench~\cite{wang2026aligningagentsplanningbenchmark}
& LLM
& Trajectory pairs
& $\checkmark$
& $\times$
& $\times$ \\

OS-Themis / OGRBench~\cite{li2026osthemisscalablecriticframework}
& LLM
& GUI trajectories
& $\times$
& $\times$
& $\times$ \\

AJ-Bench~\cite{shi2026ajbenchbenchmarkingagentasajudgeenvironmentaware}
& Agent
& Agent trajectories
& $\checkmark$
& $\times$
& $\checkmark$ \\

\midrule

\multirow{2}{*}[-0.1ex]{\textbf{SkillTV-Bench (Ours)}}
& \multirow{2}{*}[-0.1ex]{\textbf{Agent}}
& \textbf{SkillsBench execution}
& \multirow{2}{*}[-0.1ex]{\textbf{$\checkmark$}}
& \multirow{2}{*}[-0.1ex]{\textbf{$\checkmark$}}
& \multirow{2}{*}[-0.1ex]{\textbf{$\checkmark$}}
\\

&
& \textbf{trajectories}
&
&
& 
\\


\bottomrule
\end{tabular}

\parbox{0.95\linewidth}{
\emph{Note:}
$\checkmark$ indicates that a benchmark supports the corresponding property, whereas $\times$ indicates that it does not.
\emph{Skill-Aware} denotes whether the benchmark exposes the task-time skills and support explicit evaluation of their usage within an execution trajectory.
\emph{Agentic Interaction} denotes whether the judge can actively inspect evidence through an external environment, including environment states, screenshots, files, databases, tools, or final artifacts.
}
\caption{Comparison between \dataset{}(Ours) and Existing Judge Benchmarks}
\label{tab:benchmark_comparison}
\end{table*}

\section{Introduction}
LLM agents have evolved from single-turn response generators into interactive systems that can plan, invoke tools, manipulate files, and produce final artifacts~\citep{yao2023reactsynergizingreasoningacting,schick2023toolformerlanguagemodelsteach,drouin2024workarenacapablewebagents,yao2024taubenchbenchmarktoolagentuserinteraction}.
This shift also changes the object of agent evaluation, where evaluation must determine not only whether the final output appears plausible, but also whether the complete execution genuinely satisfies the essential task requirements~\citep{lu2025agentrewardbenchevaluatingautomaticevaluations,he2025trajectbenchatrajectoryawarebenchmarkevaluating}. 
Reliable judges and verifiers, therefore, become essential components of complex agent systems.

However, existing LLM judges remain unreliable on long-horizon trajectories.
Evidence of task success is distributed across tool calls, execution logs, generated files, and final artifacts. 
Judges may over-trust polished outputs, overlook hidden process failures, or fail to inspect the required external evidence~\citep{
shi2026ajbenchbenchmarkingagentasajudgeenvironmentaware,
barke2026agentrxdiagnosingaiagent}.
Rubrics define the conditions for success, but they rarely provide an operational procedure.
They do not specify what to inspect, where to find relevant evidence, or when that evidence is sufficient for a verdict.~\citep{wan2026inferencetimescalingverificationselfevolving,shi2026ajbenchbenchmarkingagentasajudgeenvironmentaware}

This problem is harder for skill-augmented agents. 
Skills encode task-relevant procedural knowledge, including tool choices, operational workflows, critical constraints, and artifact specifications~\citep{li2026skillsbenchbenchmarkingagentskills,
chen2026skilljurormeasuringagentskill,
ding2026agentskillevaluationevolution}.
Many of these requirements cannot be recovered from the final artifact alone. 
For a judge, skills therefore provide important verification context.
They indicate which failure modes to investigate, what evidence should exist, and where that evidence can be found.
Without this context, a judge may observe an execution trajectory but still misinterpret the consequences of particular actions or omissions.

Existing benchmarks do not adequately evaluate this capability.~\citep{tan2025judgebenchbenchmarkevaluatingllmbased,
lu2025agentrewardbenchevaluatingautomaticevaluations,
wang2026aligningagentsplanningbenchmark,
shi2026ajbenchbenchmarkingagentasajudgeenvironmentaware}
Many focus on final responses, response pairs, or plans rather than complete agent executions.
Benchmarks that include trajectories often present them as static text instead of directly inspectable execution packages.
They also rarely expose the skills available to the task agent, preventing direct assessment of whether a judge can use procedural context to plan inspections, locate evidence, and interpret agent behavior.
These limitations prevent existing benchmarks from systematically revealing judge failures in skill-aware and evidence-grounded trajectory verification.

To address this gap, we introduce \dataset{}, a skill-aware trajectory verification benchmark for skill-augmented agent executions. 
It contains 681 execution cases from 50 source tasks across eleven domains.
Each case combines the original instruction, a normalized trajectory, the task-time skills, and inspectable execution artifacts. 
This design supports multi-domain, skill-aware, and evidence-grounded evaluation with agentic interaction.

We also introduce \method{}, which represents verification knowledge as a reusable external JudgeSkill.
The JudgeSkill guides an agent judge to create an inspection plan, examine case-specific evidence, record its findings, and issue an evidence-grounded verdict. 
An automated development-time loop then analyzes misjudged cases and proposes targeted revisions.
Candidate revisions are promoted through a fixed development gate, allowing the judge to improve without updating the underlying model. 
Figure~\ref{fig:overview} summarizes the benchmark gaps addressed by \dataset{} and the improvements achieved by \method{} in trajectory verification and rollout selection.

\begin{figure*}[!t]
\centering
\includegraphics[width=\textwidth]{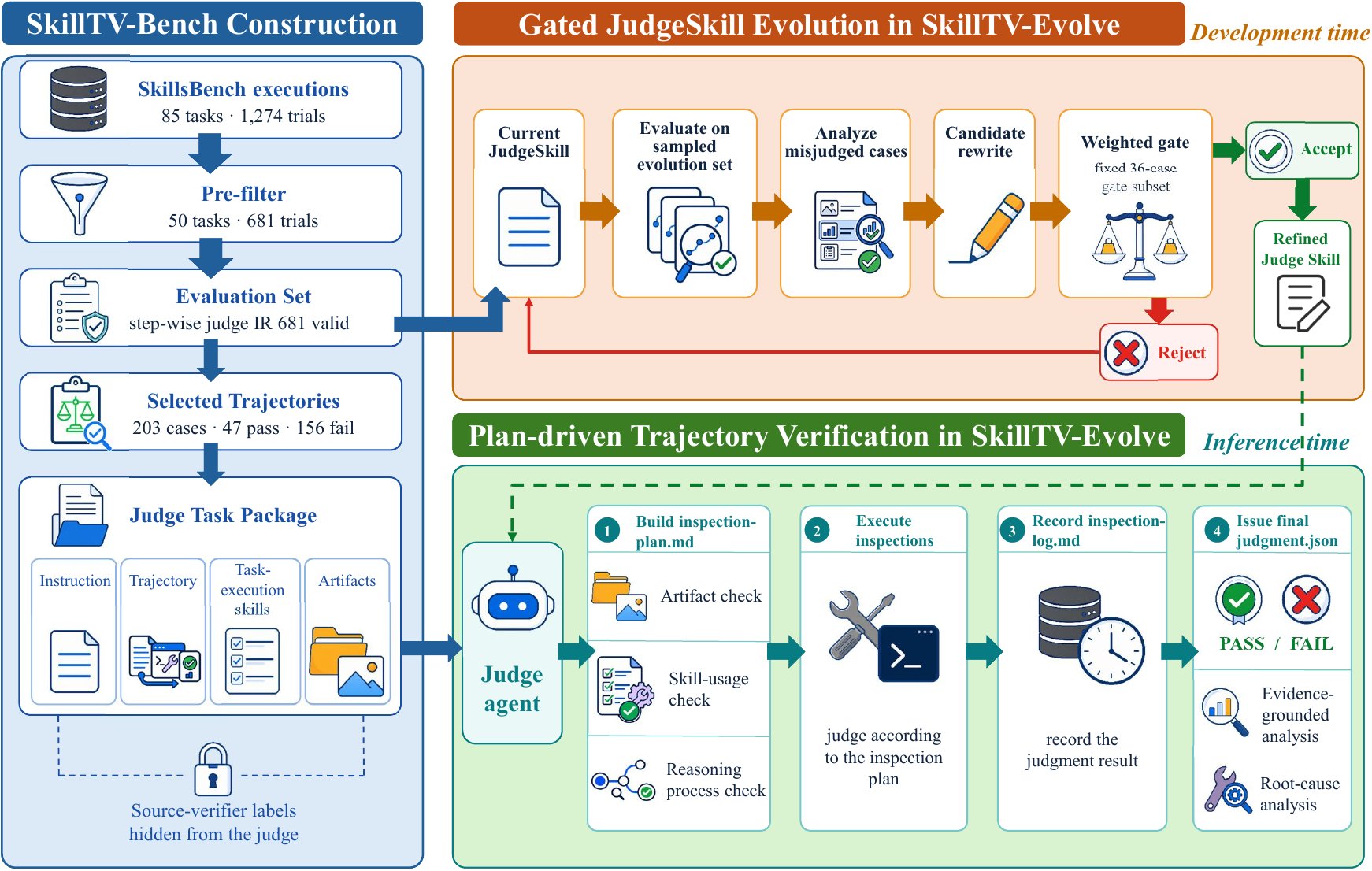}
\caption{Overview of SkillTV-Bench construction and SkillTV-Evolve, including gated JudgeSkill refinement and plan-driven, evidence-grounded trajectory verification.}
\label{fig:project_workflow}
\end{figure*}

Figure~\ref{fig:project_workflow} details the end-to-end workflow
underlying these contributions. Starting from SkillsBench executions,
we filter, normalize, and package trajectories with the task
instruction, task-time skills, and artifacts into self-contained judge
cases. On the development split, misjudged cases drive gated JudgeSkill
refinement. At inference time, the resulting JudgeSkill guides
plan-driven inspections and evidence-grounded verdicts.

Our contributions are:
\begin{enumerate}
    \item We introduce \dataset{}, a 681-case trajectory-verification benchmark spanning 50 source tasks and eleven domains, supporting multi-domain, agentic, evidence-grounded, and skill-aware evaluation.

    \item We propose \method{}, a plan-driven external JudgeSkill with an automated evolution loop that improves verification without updating model parameters.

    \item Our experiments demonstrate improved trajectory verification and agent rollout selection, validating the effectiveness of connecting benchmark-grounded evaluation with JudgeSkill evolution.
\end{enumerate}


\section{Related Work}

\paragraph{Trajectory-level agent evaluation}
LLM agents solve long-horizon tasks through planning, tool use, and environment interaction, making final-answer-only evaluation insufficient ~\citep{yao2023reactsynergizingreasoningacting,schick2023toolformerlanguagemodelsteach,drouin2024workarenacapablewebagents,yao2024taubenchbenchmarktoolagentuserinteraction}. As shown in Table~\ref{tab:benchmark_comparison}, LLM-as-a-judge and reward-model benchmarks primarily evaluate responses or preferences ~\citep{liu2023gevalnlgevaluationusing,lambert2024rewardbenchevaluatingrewardmodels,kim2024prometheus2opensource,tan2025judgebenchbenchmarkevaluatingllmbased}, while recent work extends evaluation to complete agent trajectories. AgentRewardBench studies LLM judgments of web-agent success and behavioral errors, and Plan-RewardBench evaluates pairwise preferences over tool-using trajectories ~\citep{lu2025agentrewardbenchevaluatingautomaticevaluations,wang2026aligningagentsplanningbenchmark}. Agent-as-a-Judge and AJ-Bench instead allow judge agents to inspect intermediate execution evidence or interact with external environments ~\citep{zhuge2024agentasajudgeevaluateagentsagents,shi2026ajbenchbenchmarkingagentasajudgeenvironmentaware}. TRAIL and AgentRx further study failure localization and root-cause attribution from execution traces ~\citep{deshpande2025trailtracereasoningagentic,barke2026agentrxdiagnosingaiagent}. Our setting focuses on verification of multi-domain, skill-augmented agent executions, where the judge jointly reasons over the original instruction, trajectory, task-time skills, and final artifacts.

\paragraph{Agent skills and textual evolution}
Agent skills package reusable procedural knowledge that can augment an agent without modifying model parameters. SkillsBench shows that curated skills can improve agent success ~\citep{li2026skillsbenchbenchmarkingagentskills}. Prior prompt-optimization methods use evaluated examples or execution traces to iteratively improve textual instructions ~\citep{yang2024largelanguagemodelsoptimizers,fernando2023promptbreederselfreferentialselfimprovementprompt,agrawal2026gepareflectivepromptevolution}. More recent skill-learning methods diagnose and revise procedural skills from agent experience, retaining updates according to downstream execution utility ~\citep{zhong2026skilllearnbenchbenchmarkingcontinuallearning,liu2026skillreviseimprovingllmauthoredagent}. Concurrently, Skill-RM represents heterogeneous reward evaluation as the execution of a reusable evaluation skill that orchestrates rubrics, checklists, and verifiers ~\citep{chen2026skillrmunifyingheterogeneousevaluation}. These works evaluate or improve task-agent skills; in contrast, we expose task-time skills as judge-side procedural context and evolve the JudgeSkill itself from judgment errors.

\paragraph{Verifier-guided test-time scaling}
Repeated sampling can increase the probability that a candidate pool contains a successful solution, but practical gains depend on accurately identifying that candidate ~\citep{brown2024largelanguagemonkeysscaling,snell2024scalingllmtesttimecompute}. Recent agentic test-time-scaling work studies parallel rollouts, revision, verification, and result aggregation, showing that the choice of verifier and merging strategy materially affects final performance ~\citep{zhu2025scalingtesttimecomputellmagents}. DeepVerifier uses failure-derived rubrics to evaluate and iteratively revise deep-research outputs at inference time ~\citep{wan2026inferencetimescalingverificationselfevolving}. In contrast, our approach uses verification to select among stored agent-execution trajectories without revising or rerunning the task agent.

\section{Method}

\subsection{Skill-Aware Trajectory-Level Verification}

We study verification of a completed execution rather than evaluation of its final response in isolation. A judge case is
\begin{equation}
    z_i = (x_i,y_i), \qquad x_i=(I_i,\tau_i,S_i,A_i),
\end{equation}
where $I_i$ is the original task instruction, $\tau_i$ is the normalized execution trajectory, $S_i$ is the set of skills available to the task agent, $A_i$ is the set of execution artifacts, and $y_i\in\{0,1\}$ is the source-verifier label, where 0 denotes failure and 1 denotes success. At evaluation time, the judge observes only $x_i$ and returns a verdict $\hat{y}_i$ and an evidence-grounded rationale $R_i$,
\begin{equation}
    (\hat{y}_i,R_i)=f_{\mathrm{judge}}(I_i,\tau_i,S_i,A_i),
\end{equation}
while $y_i$ and all label-revealing verifier outputs remain hidden. This formulation makes task success a joint consistency question: the delivered artifacts must satisfy the instruction; the trajectory must not contain reasoning failures that invalidate the delivered result; and, when a task-time skill imposes correctness-critical procedures or constraints, the execution should select and follow that skill appropriately. Skill awareness does not equate success with superficial skill invocation. It requires skill use, process evidence, and final outcomes to jointly support the verdict.

\subsection{\dataset{}}

\paragraph{Benchmark overview}
\dataset{} is a trajectory-level verification benchmark derived from real skill-augmented executions in SkillsBench ~\citep{li2026skillsbenchbenchmarkingagentskills}. It contains 681 cases from 50 source tasks across eleven domains: cybersecurity (68 cases), energy (39), finance (59), healthcare (15), manufacturing (31), mathematics (14), media and content (90), natural science (105), office workflows (108), robotics (53), and software engineering (99). We partition these cases into a 478-case set for judge evolution and
a disjoint 203-case evaluation set. The split is task-disjoint: no source task contributes executions to both partitions. Accordingly, our experiments report precision, recall, and F1 together with accuracy and balanced accuracy. Each case is packaged as the observable tuple $(I,\tau,S,A)$.

\paragraph{Benchmark construction}
We construct the benchmark in five stages. First, we extract 1,274 raw trials from 85 SkillsBench tasks. Every trial contributes the result, agent trajectory, and source-verifier label. Second, we remove overlong trajectories and retain source tasks with both pass and fail executions remaining, yielding 681 trials from 50 tasks. We believe there is no need to introduce verifiers to select pass candidate if the agentic trajectories of the task are completely correct or wrong. Third, a deterministic rule-based converter maps each unstructured agent event stream into a compact step-wise representation with a fixed schema. This preserves high-signal execution evidence without exposing the judge to redundant raw logs. Fourth, we validate schema conformance, token and step limits, and cross-field consistency; all 681 normalized trials pass these checks. Finally, we select 203 cases from 14 source tasks in nine domains using domain-balanced sampling and assign the remaining 478 cases to the evolution set.

The construction step joins each normalized trajectory with the task-time skill set and artifacts to create a self-contained judge environment. The task-time skill set enables our skill-aware verification, where judges can inspect the skill utilization and orchestration in the agent execution. The original instruction of the agentic task acts as an environment map: it tells the judge what requirements should be traced into the judge environment. Consequently, \dataset{} measures evidence-grounded inspection of a rich execution package, not final-answer plausibility or simple agent trajectory only.

\subsection{\method{}}

\paragraph{Inspection-plan-driven judging}
Our method represents verification knowledge as a human-readable JudgeSkill that structures inference into three explicit products: an inspection plan, an inspection log, and an evidence-grounded judgment,
\begin{equation}
    (I,\tau,S,A) \rightarrow P \rightarrow L \rightarrow (\hat{y}, R),
\end{equation}
where $P$ specifies what to check, where to find the relevant evidence, and what would count as sufficient support or refutation; $L$ records the checks actually executed and their findings; and $(\hat{y},R)$ is the verdict with its diagnostic rationale. The judge implements artifact, skill-usage, and reasoning-process checks on the instruction and case package and weighs any conflicting evidence before reaching a decision. Unlike a rubric, which primarily states desired properties, the inspection plan operationalizes them into evidence-seeking actions. A pass requires the log to contain sufficient positive evidence and no supported task-critical failure, such as a missing required artifact, unresolved runtime error, schema violation, omitted verification step, or skill-use error that invalidates the result. To supply compact failure priors, we synthesize three judge-facing references from prior empirical analyses paper of agent failures: first concerns failures visible in artifacts or final environment state, second focuses on failures visible in reasoning and intermediate execution, and a third covers failures in agentic skill utilization and orchestration.~\citep{deshpande2025trailtracereasoningagentic,ma2026demystifyinglifecyclefailuresplatformorchestrated,shah2026characterizingfaultsagenticai,zhu2025llmagentsfaillearn,barke2026agentrxdiagnosingaiagent,song2025aegistaxonomyoptimizationsovercoming,rahardja2025agentsfixagentissues,zhu2026empiricalstudybugsmodern,lu2025exploringautonomousagentscloser}. These references guide judge reasoning process, but the case-specific plan and observed evidence determine the verdict.

\begin{table*}[t]
\centering
\begin{tabular}{lccccc}
\toprule
Method & Precision & Recall & F1 & Accuracy & Bal. Acc. \\
\midrule
GPT-5.2 
    & 0.216 & 0.468 & 0.295 & 0.483 & 0.478 \\
Claude Sonnet 4.6 
    & 0.258 & \textbf{0.915} & 0.402 & 0.370 & 0.560 \\
Gemini 3.1 Pro Preview 
    & 0.274 & 0.851 & 0.415 & 0.443 & 0.586 \\
Multi-judge
    & 0.269 & 0.851 & 0.408 & 0.429 & 0.576 \\
Rubric-based Claude Sonnet 4.6
    & 0.250 & 0.660 & 0.363 & 0.463 & 0.532 \\
Agent judge w/o JudgeSkill
    & 0.266 & 0.809 & 0.400 & 0.438 & 0.568 \\
\midrule
\textbf{Agent judge w/ refined JudgeSkill(Ours)}
    & \textbf{0.324} & 0.723 & \textbf{0.447}
    & \textbf{0.586} & \textbf{0.634} \\
\bottomrule
\end{tabular}
\caption{Trajectory-verification results on the \dataset{} evaluation set. The first six rows characterize existing judging approaches (RQ1), while the final row reports the refined \method{} judge (RQ2).}
\label{tab:main_results}
\end{table*}

\paragraph{Automated JudgeSkill evolution}
We improve the external skill text without updating model parameters. The evolution data are drawn from a pool of 478 trajectories disjoint from the evaluation set. In each round, the current judge agent inferences on sampled development executions; misjudged cases are then partitioned into reflection bundles for three refinement goals: false-pass hardening, false-fail relaxation, or balanced calibration. Each bundle contains the judge's plan, log, verdict, and development-only source-verifier evidence explaining the true label. Because full traces are long, the system compresses each bundle and a shared train-signal bundle into case-level rewrite briefs. An optimizer receives the current skill, its reference documents, one strategy-specific brief, and the shared brief, and proposes a revised skill. Source-verifier evidence is used only to generate these offline rewrite signals and is never available during benchmark judgment.

Candidate generation and version selection are separated. Every proposed rewrite is evaluated on a fixed 36-case development gate that preserves all 14 task families, contains 27 fail and 9 pass cases, and matches the full-set baseline behavior. A candidate replaces the incumbent only if it improves
\begin{equation}
G = 0.5\cdot \mathrm{BalAcc}+0.3\cdot \mathrm{Recall}_{\mathrm{fail}}+0.2\cdot \mathrm{Precision}_{\mathrm{pass}}.
\end{equation}
The gate emphasizes rejection of failed executions while preserving calibrated acceptance of genuine successes. The resulting loop of evaluation, reflection, rewriting, gating, and promotion turns JudgeSkill development into an auditable optimization process.

\section{Evaluation}

We conduct experiments to answer the following research questions:
\begin{description}
    \item[\textbf{RQ1}] Does \dataset{} reveal systematic weaknesses in \hspace*{1.25em}existing trajectory judges?

    \item[\textbf{RQ2}] Can \method{} use the errors exposed by the \hspace*{1.25em}benchmark to improve trajectory verification?

    \item[\textbf{RQ3}] Do improvements in verification translate into better \hspace*{1.25em}rollout selection?

\end{description}

\subsection{Experimental Setup}

\paragraph{Evaluation set and metrics}
All judges are evaluated on a held-out set of 203 cases spanning 14 source tasks and nine domains, selected from the full \dataset{} collection of 681 trajectories through domain-balanced sampling. We report pass-class precision, recall, and F1 to measure the reliability and coverage of accepted executions, overall accuracy, and balanced accuracy to account equally for performance on successful and failed trajectories. The McNemar's test is used for paired comparisons between agent-judge variants.

\paragraph{Evaluation infrastructure}
We evaluate GPT-5.2, Claude Sonnet 4.6, and Gemini 3.1 Pro Preview as direct LLM judges ~\citep{openai2025gpt52,anthropic2026sonnet46,gemini2026gemini31pro}. The rubric-based judge uses Claude Sonnet 4.6. All agent-as-a-judge variants use Claude Sonnet 4.6 as the backbone model and Claude Code v2.1.19 as the agent harness. Judge agents run in containerized Harbor environments ~\citep{HarborFrameworkTeam2026,merrill2026terminalbenchbenchmarkingagentshard}, where the task instruction, normalized trajectory, task-time skills, and final artifacts are mounted for inspection.

\subsection{RQ1: Benchmark Insights}

\paragraph{\dataset{} remains challenging for existing judges}
Table~\ref{tab:main_results} shows that \dataset{} presents a substantial challenge even to strong contemporary judges. The three frontier direct judges achieve balanced accuracies of only 0.478--0.586. Combining their predictions through majority voting yields a balanced accuracy of 0.576, while task-specific rubric generation reaches only 0.532. Providing an agent judge with tools and direct access to the execution environment is also insufficient by itself: without a dedicated JudgeSkill, the agent obtains a balanced accuracy of 0.568. Thus, model scale, model ensembling, static rubrics, and agentic environment inspection do not individually resolve the verification problem defined by \dataset{}.

\paragraph{Existing judges frequently accept plausible failures}
The dominant error pattern is not simply an inability to recognize successful executions. Instead, existing judges frequently accept failed trajectories that appear plausible. Claude Sonnet 4.6, for example, achieves the highest recall, but its precision and overall accuracy are quite low. Gemini 3.1 Pro Preview and the multi-judge baseline also exhibit a similar pattern.

This precision--recall imbalance indicates that the main challenge exposed by \dataset{} is distinguishing genuinely successful executions from plausible-but-failed ones. This calls for stronger judge capabilities in identifying critical agentic failures. 

\subsection{RQ2: Benchmark-Grounded Improvement}

\paragraph{Benchmark results can be converted into verification knowledge}
Beyond revealing verification failures, \dataset{} provides development-time signals for improving judge behavior. \method{} analyzes false accepts and false rejects on development cases that are disjoint from the evaluation set. It then converts recurring judgment failures into revisions of an explicit JudgeSkill, which specifies how the judge should plan inspections, gather evidence, and reconcile conflicting signals. Candidate revisions are promoted only through the fixed development gate; the evaluation labels are never used for skill evolution.

As shown in Table~\ref{tab:main_results}, the refined JudgeSkill increases the same agent judge's balanced accuracy from 0.568 to 0.634 and its overall accuracy by 14.8 percentage points. This paired accuracy improvement is significant under McNemar's test ($p<10^{-4}$), and the refined skill also significantly outperforms the original JudgeSkill ($p<0.01$).

\paragraph{Improvements across domains}
Table~\ref{tab:domain_results} shows that the gain is broad, although heterogeneous across domains. The refined JudgeSkill improves accuracy in seven of the nine evaluation domains and does not reduce accuracy in any domain. The largest improvement occurs in media and content (+36.7 percentage points), followed by cybersecurity (+26.7 pp), manufacturing (+23.1 pp), finance (+21.4 pp), and energy (+20.0 pp). Software engineering also improves substantially (+16.7 pp). Natural science improves by 3.3 percentage points, while office workflows and robotics remain unchanged. The aggregate improvement is therefore not attributable to a single domain.

\begin{table}[!t]
\centering
\resizebox{\linewidth}{!}{
\begin{tabular}{lcccc}
\toprule
Domain & Cases & No JudgeSkill & Refined & $\Delta$(pp) \\
\midrule
Cybersecurity       & 15 & 40.0\% & 66.7\% & +26.7 \\
Energy              & 15 & 66.7\% & 86.7\% & +20.0 \\
Finance             & 14 & 50.0\% & 71.4\% & +21.4 \\
Manufacturing       & 13 & 30.8\% & 53.8\% & +23.1 \\
Media and content   & 30 & 33.3\% & 70.0\% & +36.7 \\
Natural science     & 30 & 50.0\% & 53.3\% & +3.3 \\
Office workflows    & 27 & 48.1\% & 48.1\% & +0.0 \\
Robotics            & 29 & 37.9\% & 37.9\% & +0.0 \\
Software engineering& 30 & 43.3\% & 60.0\% & +16.7 \\
\midrule
Overall             & 203 & 43.8\% & 58.6\% & +14.8 \\
\bottomrule
\end{tabular}
}
\caption{Domain-level comparison of the agent judge’s performance with and without the refined JudgeSkill.}
\label{tab:domain_results}
\end{table}

\paragraph{Progress across JudgeSkill versions}
Table~\ref{tab:skill_evolution} traces the effect of benchmark-grounded evolution. From v1 to v4, accuracy on the fixed development gate increases by 5.5 percentage points, while the gate score rises by 3.5 percentage points. Accuracy on the full evaluation set improves monotonically from 50.75\% to 58.62\%.

Balanced accuracy on the development gate peaks at v3, whereas v4 achieves the highest gate score and full set accuracy. The progression across versions shows that the benchmark does not merely rank fixed judges: its development cases provide actionable error signals that can be externalized as reusable verification knowledge.

\begin{table}[!t]
\centering

\resizebox{\linewidth}{!}{
\begin{tabular}{@{}l@{\hspace{4pt}}ccc@{\hspace{2pt}}c@{}}
\toprule
Method & Acc.(36-case) & Bal. Acc. & Gate $G$ & Acc.(full) \\
\midrule
Agent judge w/ JudgeSkill v1
    & 52.8\% & 57.4\% & 49.2\% & 50.8\% \\
Agent judge w/ JudgeSkill v2
    & 52.8\% & 61.1\% & 50.3\% & 54.2\% \\
Agent judge w/ JudgeSkill v3
    & 55.6\% & \textbf{63.0\%} & 52.6\% & 55.7\% \\
Agent judge w/ JudgeSkill v4
    & \textbf{58.3\%}
    & 57.4\% & \textbf{52.7\%} & \textbf{58.6\%} \\
\bottomrule
\end{tabular}
}
\caption{JudgeSkill versions on the fixed 36-case development gate and the full \dataset{} evaluation set.}
\label{tab:skill_evolution}
\end{table}

\subsection{RQ3: Downstream Rollout Selection}

\paragraph{Offline rollout-selection protocol}
To test the downstream utility of verification, we simulate test-time scaling where an agent generates multiple rollouts and a verifier selects the trajectory to deliver, using stored trajectory pools rather than rerunning the task agents. For each of the 14 source tasks, each rollout budget $n\in\{1,\ldots,10\}$, and eight paired repeats, we sample $n$ trajectories without replacement and apply every judging method to the same sampled pool. We select uniformly from the trajectories predicted as pass and consult the hidden source-verifier label to determine whether the selected trajectory succeeds. We report the success rate macro-averaged across tasks and then averaged across repeats. At $n=1$, no selection is possible, so all methods recover the underlying single-rollout success rate of 22.9\%; results at larger $n$ measure whether verifier-guided selection can convert additional rollouts into higher task success.

\begin{figure}[!t]
\centering
\includegraphics[width=\columnwidth]{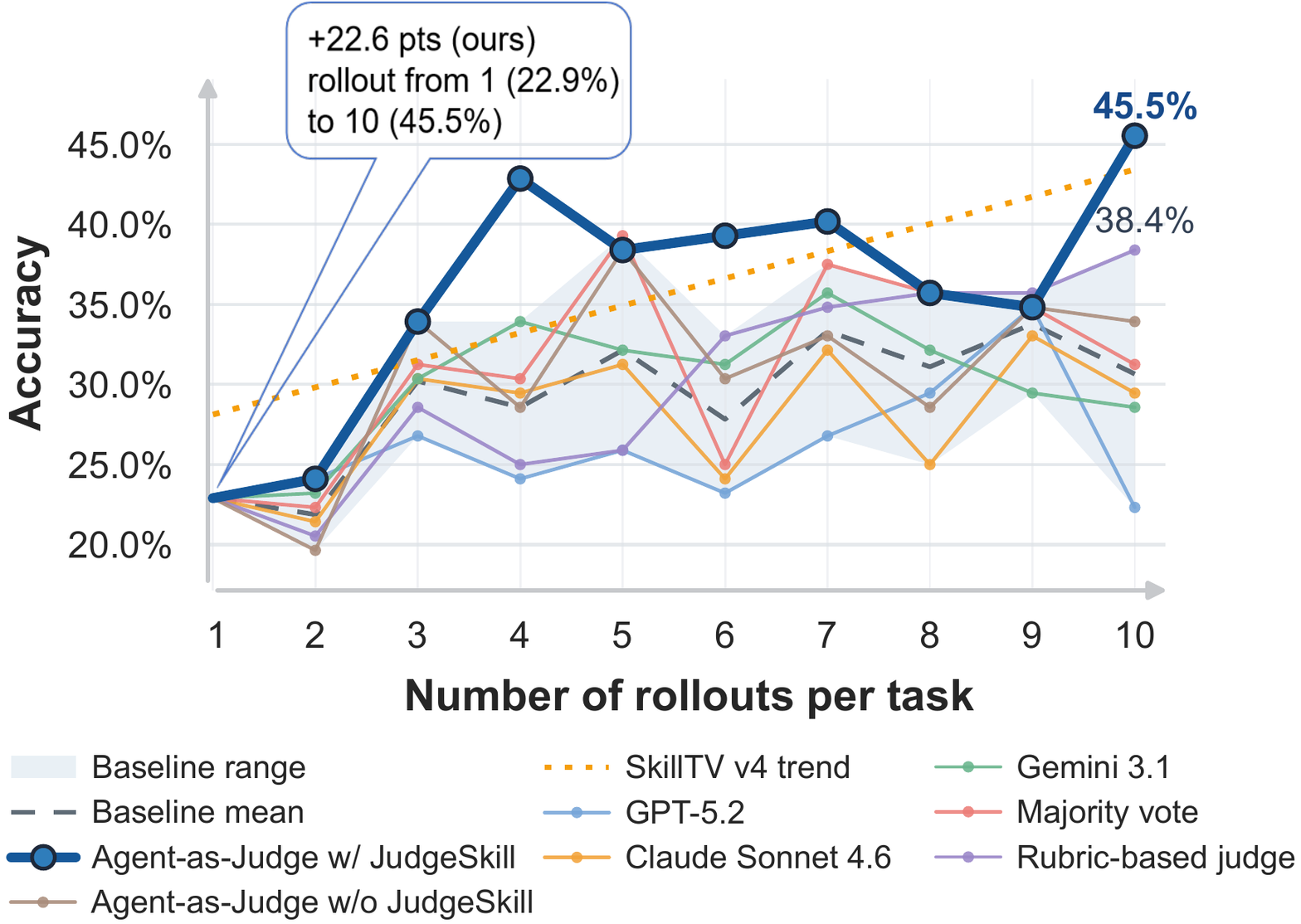}
\caption{Offline verifier-guided rollout selection over 14 source tasks.
The shaded region and dashed gray line denote the range and mean of the baselines, respectively; the orange dotted line shows the overall trend of JudgeSkill v4.}
\label{fig:test_time_scaling}
\end{figure}

\paragraph{Better verification yields higher selected-trajectory success}
Figure~\ref{fig:test_time_scaling} illustrates how selected-trajectory success changes with the rollout budget under verifier-guided selection, while Table~\ref{tab:rollout-summary} reports the exact results at ten rollouts. The refined JudgeSkill shows a clear overall improvement as more rollouts become available. At ten rollouts, it achieves 45.5\% success, a gain of 22.6 percentage points over the single-rollout setting, exceeding the strongest baseline, rubric-based Claude Sonnet 4.6, by 7.1 points and the agent judge without a JudgeSkill by 11.6 points.

In contrast, GPT-5.2 falls slightly below the single-rollout baseline at ten rollouts, while majority voting and the agent judge without a JudgeSkill peak at five rollouts before declining. Increasing candidate coverage therefore does not guarantee higher delivered success: Additional candidates are beneficial rather than burdensome only when the verifier can distinguish successful trajectories from plausible failures.

\paragraph{False acceptances directly impair candidate selection}
The false-accept problem revealed by \dataset{} has a direct downstream consequence. Under our filtering protocol, the selected trajectory is sampled uniformly from all candidates predicted as pass. Every failed trajectory that is incorrectly accepted therefore dilutes the probability of selecting a genuinely successful execution. Reducing false accepts from 105 to 71 allows the refined JudgeSkill to preserve a substantially cleaner predicted-pass set, explaining why its verification improvement translates into higher selected-trajectory success.

\begin{figure}[!t]
\centering
\includegraphics[width=0.98\columnwidth]{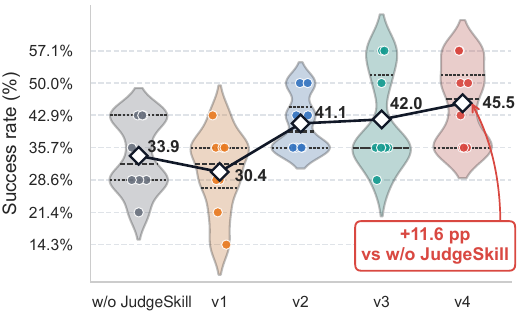}
\caption{Agent-as-a-judge comparison at ten rollouts over eight paired repeats. Success improves across evolved JudgeSkill versions, with v4 achieving the highest mean selected-trajectory success.}
\label{fig:rollout10_agent_evolution}
\end{figure}

Figure~\ref{fig:rollout10_agent_evolution} further shows that merely adding a JudgeSkill is insufficient. At ten rollouts, the original JudgeSkill v1 reaches only 30.4\%, below the 33.9\% obtained by the base agent judge. Benchmark-grounded evolution subsequently raises selected-trajectory success to 41.1\%, 42.0\%, and 45.5\% for v2, v3, and v4, respectively. These results connect the benchmark's central error pattern to practical selection performance. By reducing false acceptances, the refined JudgeSkill keeps the predicted-pass candidate set cleaner, thereby increasing the probability of selecting a genuinely successful trajectory from the rollout pool.

\begin{table}[!t]
\centering

\resizebox{\linewidth}{!}{
\begin{tabular}{lccc}
\toprule
Method & Success@10 & $\Delta$ & Best rollout \\
\midrule
GPT-5.2
    & 22.3\% & $-0.6$ & 9 \\
Claude Sonnet 4.6
    & 29.5\% & $+6.6$ & 9 \\
Gemini 3.1 Pro Preview
    & 28.6\% & $+5.7$ & 7 \\
Multi-judge
    & 31.3\% & $+8.4$ & 5 \\
Rubric-based judge
    & 38.4\% & $+15.5$ & 10 \\
Agent judge w/o JudgeSkill
    & 33.9\% & $+11.0$ & 5 \\
\midrule
Agent judge w/ JudgeSkill v1
    & 30.4\% & $+7.5$ & 9 \\
Agent judge w/ JudgeSkill v2
    & 41.1\% & $+18.2$ & 10 \\
Agent judge w/ JudgeSkill v3
    & 42.0\% & $+19.1$ & 10 \\
Agent judge w/ JudgeSkill v4
    & \textbf{45.5\%} & \textbf{$+22.6$} & \textbf{10} \\
\bottomrule
\end{tabular}
}
\caption{Offline rollout-pool selection at ten rollouts. Success@10 is the mean selected-trajectory success over eight paired repeats, and $\Delta$ is the percentage-point difference from the common 22.9\% single-rollout baseline.}
\label{tab:rollout-summary}
\end{table}

\section{Discussion}

Our results suggest that trajectory verification should be viewed not merely as passive evaluation, but as the mechanism that converts diverse agent rollouts into reliable task outcomes. SkillTV-Bench makes this problem measurable by requiring judges to reconcile task instructions, execution processes, skill-use requirements, and inspectable artifacts, while SkillTV-Evolve shows that the required verification knowledge can be externalized as an explicit, auditable procedure and refined from judgment errors without updating model parameters. The experiments show that skill-aware trajectory verification is both systematically improvable and practically consequential: evolving an explicit JudgeSkill yields consistent gains in the verification capability, and these gains eventually translate into higher selected-trajectory success. 

A key limitation, however, is that the current loop improves only the judge. Although SkillTV-Evolve turns benchmark errors into reusable verification knowledge, the evidence-grounded feedback of our judge is not yet returned to task agents to produce better task executions or generate higher-quality trajectories. Consequently, the present study demonstrates improved verification and offline selection over existing rollout pools, but not end-to-end task-agent improvement or automatic benchmark expansion. Closing this gap would turn the verifier from a candidate filter into a learning signal, allowing validated new trajectories to expand SkillTV-Bench and support a complete, continuously improving data flywheel.

\section{Conclusion}
We introduced \dataset{} and \method{} to address a central question in agent evaluation: how can we reliably verify whether a long-horizon agent execution truly satisfies its task? \dataset{} reveals systematic weakness in existing trajectory judges, while \method{} provides a continuously evolving verification mechanism. Experiments prove the evolved skill not only improves verification but also translates these gains into higher selected-trajectory success. 

Looking ahead, a natural next step is to connect \dataset{} and \method{} in a closed loop between trajectory verification and task execution. While our methods mainly focus on using judgment errors to improve the JudgeSkill, future work could return the judge's evidence-grounded feedback to task agents, helping them generate higher-quality trajectories. After appropriate selection, these new trajectories could expand SkillTV-Bench and reveal new verification challenges, which would in turn drive further JudgeSkill evolution. Such a feedback loop would allow task agents, judges, and the benchmark to improve together, turning trajectory verification from a final evaluation step into a reusable learning signal for continuously evolving agent development.

\bibliography{main}

\end{document}